\documentclass[11pt]{article}

\usepackage{acl}

\usepackage{times}
\usepackage{latexsym}

\usepackage[T1]{fontenc}

\usepackage[utf8]{inputenc}

\usepackage{microtype}

\usepackage{inconsolata}

\usepackage{graphicx}

\title{\textbf{ChitraMiti: Benchmarking Visual Grounding and Modality Reliance in
       Bengali Geometric Reasoning}}

\author{
  Khan Raiyan Ibne Reza, \enspace Sanjana Aktar Maria, \enspace Sumaiya Tabassum Nimi, \enspace Md Adnan Arefeen \\
  \textmd{North South University} \\[2.5pt]
  \href{https://huggingface.co/datasets/RaiyanKhaan/ChitraMiti}{\color{blue}\textmd{HuggingFace/Datasets/RaiyanKhaan/ChitraMiti}}
}
\date{}

\graphicspath{{figures/}}
\usepackage{booktabs}
\usepackage{multirow}
\usepackage{array}
\usepackage{amsmath,amssymb}
\usepackage{float}
\usepackage[skins,breakable]{tcolorbox}
\usepackage{fontspec}
\usepackage{polyglossia}
\setdefaultlanguage{english}
\setotherlanguage{bengali}
\IfFontExistsTF{Times New Roman}{%
  \setmainfont{Times New Roman}
}{%
  \IfFontExistsTF{texgyretermes-regular.otf}{%
    \setmainfont{texgyretermes}[
      Path=./,
      Extension=.otf,
      UprightFont=*-regular,
      BoldFont=*-bold,
      ItalicFont=*-italic,
      BoldItalicFont=*-bolditalic
    ]
  }{%
    \setmainfont{TeX Gyre Termes}
  }
}

\IfFontExistsTF{NotoSansBengali-Regular.ttf}{%
  \newfontfamily\bengalifont{NotoSansBengali-Regular.ttf}[Path=./, Script=Bengali]
  \newfontfamily\bengalifonttt{NotoSansBengali-Regular.ttf}[Path=./, Script=Bengali]
}{%
  \newfontfamily\bengalifont{Noto Sans Bengali}[Script=Bengali]
}

\definecolor{promptbg}{gray}{0.93}

\begin{document}

\maketitle

\begin{abstract}
Evaluation of vision-language models (VLMs) for multimodal mathematical reasoning remains limited for low-resource languages and for geometry problems that require reading a diagram and a question together. We introduce \textbf{ChitraMiti-12.8k}, a synthetic benchmark of 12,874 Bengali planar geometry problems paired with structured 15-attribute descriptions, and \textbf{NCTB-500}, a complementary set of 500 diagrams manually extracted from Bengali school textbooks. Using a three-phase protocol that separates diagram-only, diagram-plus-description, and description-only inputs, we show across five open-weight and closed-source VLMs that description-only performance is statistically indistinguishable from diagram-plus-description performance, establishing structured descriptions as a sufficient textual proxy for controlled evaluation. Despite this, models remain poor at cross-modal verification, frequently misled by a swapped spatial relation even when they answer the unmodified item correctly. We further evaluate supervised adaptation on ChitraMiti-12.8k, finding that fine-tuning improves performance on both ChitraMiti-1k and NCTB-500, although a substantial gap to the strongest zero-shot model remains. Together, ChitraMiti-12.8k, NCTB-500, and our evaluation protocol offer a standardized way to study Bengali multimodal geometry reasoning and, more broadly, whether VLMs actually check their text against what they see.
\end{abstract}

\section{Introduction}
\label{sec:introduction}

Multimodal mathematical reasoning benchmarks typically pair geometric diagrams with natural language questions to evaluate vision-language models (VLMs) \citep{lu2024mathvista}. Evaluating progress in low-resource languages is more difficult because large-scale multimodal benchmarks are scarce, existing Bengali math resources remain text-only \citep{hasan2021xl, bhattacharjee2023banglanlg, prama2025banglamath}.

We address this gap with \textbf{ChitraMiti-12.8k}, a large-scale synthetic Bengali planar geometry benchmark built using a shared annotation pipeline and structured-description schema. ChitraMiti-12.8k extends the DeepVision-103k corpus \citep{sun2026deepvision} to Bengali, producing 12,874 geometry problems with structured Bengali descriptions. To test whether results on synthetic diagrams generalize to classroom material, we further construct \textbf{NCTB-500}, an evaluation set of 500 manually extracted geometry diagrams from NCTB textbooks. The diagrams are annotated under the same schema, enabling comparison between synthetic and textbook diagrams.

We evaluate models under three input conditions: diagram-only (Phase A), diagram-plus-description (Phase B), and description-only (Phase C). Statistical equivalence between Phase B and Phase C indicates that the textual representation preserves sufficient information for the evaluated tasks. We use this protocol to test whether models verify textual spatial relations against the diagram, through adversarial spatial perturbations and ablations that separate relational from numeric information.



The main contributions of this work are as follows:
\begin{itemize}
\item We present \textbf{ChitraMiti-12.8k}: a large-scale synthetic benchmark of 12,874 Bengali planar geometry problems, each paired with a verified 15-attribute structured description ($\kappa=0.81$ inter-annotator agreement).
\item We build \textbf{NCTB-500}: a 500-diagram-question pair dataset manually extracted from official Bengali NCTB textbooks (Classes 6--10), serving as a test for real-world generalization that helps validate conclusions drawn from synthetic benchmarks.
\item A three-phase evaluation protocol (diagram-only, diagram-plus-description, description-only) is statistically validated using TOST equivalence testing across five VLMs and both benchmarks.
\item Our experiments demonstrate that Vision-Language Models (VLMs) rely heavily on relational text, showing ablation gaps exceeding 12 points. They also struggle to identify contradictions between text and diagrams, with accuracy dropping by 2.2\% to 12.0\% during adversarial spatial-relation swaps. This indicates that high performance on aligned inputs does not guarantee effective visual grounding.
\item Fine-tuning on Qwen-3.5 4B has improved the in-distribution accuracy from 12.7\% to 34.2\%, and the test performance on NCTB-500 increased from 9.6\% to 23.6\%. 
\end{itemize}
\section{Related Work}
\label{sec:related}

\paragraph{Multimodal mathematical reasoning benchmarks.} Existing benchmarks assess VLMs on tasks requiring joint interpretation of diagrams and text, spanning planar geometry, abstract diagrams, charts, and general visual mathematics \citep{trinh2024solving, lu2024mathvista, seo2015solving, chen2021geoqa, lu2021iconqa, zhang2024mathverse, wang2024measuring}. Most benchmarks are built in English, rely primarily on synthetic or template-generated diagrams, and few provide a standardized evaluation protocol. Our work follows this benchmark tradition by pairing a synthetic dataset with a textbook counterpart, while introducing a standardized protocol for evaluating visual diagram contributions under controlled input conditions.

\paragraph{Low-resource and Bengali NLP.} Despite its large population of speakers, Bengali remains underserved in multimodal NLP, with existing resources focusing mainly on text-only language understanding and generation \citep{kakwani2020indicnlpsuite, hasan2021xl, bhattacharjee2022banglabert, bhattacharjee2023banglanlg}. Recent Bengali mathematical benchmarks, such as BanglaMATH \citep{prama2025banglamath}, evaluate text-based mathematical reasoning but do not include geometric diagrams or cross-modal consistency evaluation. To our knowledge, no prior benchmark evaluates Bengali multimodal mathematical reasoning across synthetic and textbook diagrams under a shared annotation schema. ChitraMiti-12.8k and NCTB-500 are specifically constructed to address this gap (Section~\ref{sec:datasets}).

\paragraph{Shortcut learning and diagnostic evaluation in VLMs.} A separate line of work shows that multimodal models often rely on textual shortcuts instead of fully using visual evidence \citep{geirhos2020shortcut}. Similar behavior has been reported in visual question answering \citep{agrawal2016analyzing, goyal2017making}, visual commonsense reasoning \citep{zellers2019recognition}, chart understanding \citep{masry2022chartqa}, and MLLM perception studies \citep{tong2024eyes}.These studies typically remove visual input to measure how much models depend on text, but they do not test whether a model can detect contradictions between text and image. More recent work on multimodal inconsistency and knowledge-conflict evaluation addresses this question by examining how VLMs behave when textual and visual evidence disagree \citep{zhu2024unraveling}. This line of work has not focused on Bengali geometric reasoning with controlled structured spatial descriptions.
\section{Datasets}
\label{sec:datasets}

\begin{figure}[t]
\centering
\includegraphics[width=\linewidth]{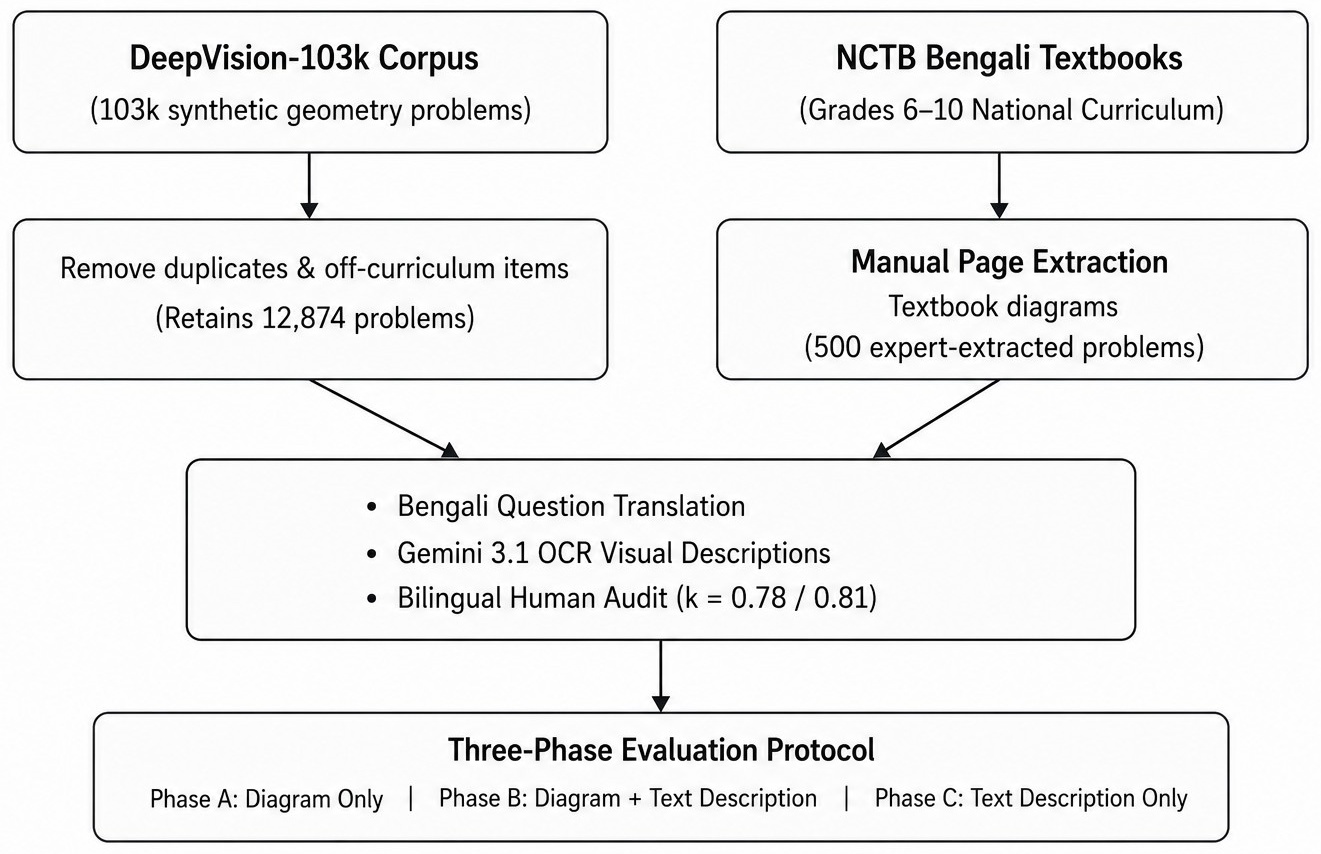}
\caption{The matched construction and evaluation pipeline for ChitraMiti-12.8k and NCTB-500.}
\label{fig:pipeline}
\vspace{-0.5em}
\end{figure}

We construct two Bengali planar geometry benchmarks through one shared pipeline: source selection, geometry topic filtering, description generation, and human verification (Figure~\ref{fig:pipeline}). For ChitraMiti-12.8k, we translate English questions into Bengali; for NCTB-500, we retain the original Bengali textbook questions. In both cases we generate structured Bengali descriptions and verify them against the source image. Both benchmarks share the same 15-attribute schema (Appendix~\ref{app:schema}).

\subsection{ChitraMiti-12.8k}
\label{ssec:chitramiti}

\paragraph{Source selection and filtering.} We use DeepVision-103k \citep{sun2026deepvision}, a large publicly available synthetic planar geometry corpus with paired diagrams and questions which contains diagrams across multiple domains. We first extract the 52,000 items explicitly annotated as planar geometry, then draw a uniform random sample of 26,000 items to bound downstream processing costs. Finally, we match these items against textbook geometry keywords and remove any residual 3D or Chinese-labeled diagrams, yielding the 12,874-item dataset (full stage-by-stage counts in Appendix~\ref{app:extended_stats}, Table~\ref{tab:pipeline}). The proportion of the categories follow the distribution of the filtered DeepVision-103k subset rather than artificial balancing. Composite geometry makes up 53.8\% of the dataset, while the remaining categories contain enough examples for category-level analysis (Table~\ref{tab:extended_stats}).

\begin{table}[t]
\centering\small
\begin{tabular}{@{}lr@{}}
\toprule
\textbf{Property} & \textbf{Value} \\
\midrule
Total samples & 12{,}874 \\
Source corpus & DeepVision-103k \\
Category: Triangles & 13.4\% (1{,}720) \\
Category: Circles & 14.5\% (1{,}861) \\
Category: Coord.\ geom. & 10.3\% (1{,}325) \\
Category: Quadrilaterals & 8.1\% (1{,}037) \\
Category: Composite geom. & 53.8\% (6{,}931) \\
Avg.\ Bengali question length & 31.9 words \\
Avg.\ description length & 66.2 words \\
\bottomrule
\end{tabular}
\caption{Dataset statistics for ChitraMiti-12.8k.}
\label{tab:extended_stats}
\end{table}

\paragraph{Translation, description generation, and verification.} We use Gemini~3.1~Flash~Lite \citep{gemini2026flashlite} to translate the questions and generate structured descriptions using a fixed prompt (Appendix~\ref{app:prompts}, \ref{app:schema}). Numerical values, variable names, and mathematical expressions are preserved without localization; only the surrounding natural-language text is translated. The structured descriptions follow a fixed 15-attribute schema (Appendix~\ref{app:schema}), and the diagram images are left unchanged. Three bilingual annotators with backgrounds in mathematics education then audited a stratified sample of 200 items, sampled proportionally across geometry categories and description-length quartiles, checking description accuracy against the diagram and translation fidelity against the source question. The audit yielded a 3.0\% correction rate (6 of 200), with inter-annotator agreement of $\kappa=0.81$ for descriptions and $\kappa=0.78$ for translations, both in the substantial-agreement range (full workflow and correction rate in Appendix~\ref{app:extended_stats}, Table~\ref{tab:annotation}).

\paragraph{Splits and integrity.} We partition ChitraMiti-12.8k into 11,231 training, 643 validation, and 1,000 test items (constructed from held-out items with verified answer extraction). Train and test items are drawn from disjoint DeepVision source IDs; no diagram appears in more than one split. Near-duplicate detection via perceptual hashing (Appendix~\ref{app:extended_stats}) confirmed fewer than 0.1\% candidate near-duplicates across splits, all subsequently removed. This produces a reproducible benchmark suitable for both zero-shot evaluation and supervised transfer experiments.

\subsection{NCTB-500}
\label{ssec:nctb500}

NCTB-500 provides 500 diagram-question pairs drawn from official Bengali-medium geometry textbooks covering Classes~6 to~10. Textbook pages were scanned and diagrams identified manually, preserving the visual fidelity of scanned classroom material rather than applying an automated OCR pipeline to diagram content. The 500-item count reflects exhaustive coverage of all planar geometry categories across five grade levels, drawn specifically from geometry chapters; extending further would require re-sampling from the same source pages, risking near-duplicate items. The difficulty distribution is 78.6\% hard, 19.0\% medium, and 2.4\% easy, with labels assigned during manual extraction based on the cognitive complexity of the geometric reasoning required (e.g., direct application of a single theorem vs.\ multi-step deduction). The composition of categories, questions, and further collection details are reported in Appendix~\ref{app:per_category} (Table~\ref{tab:nctb_dist}).

\section{Evaluation Protocol}
\label{sec:protocol}

We evaluate all models under a unified protocol designed to separate visual perception from reasoning over structured textual information. The same evaluation settings are applied to both ChitraMiti-12.8k and NCTB-500, allowing direct comparison between the synthetic and textbook-derived datasets.

\subsection{Three-Phase Evaluation}
\label{ssec:phases}

Each problem is evaluated under the three input conditions summarized in Table~\ref{tab:phases}. Phase~A provides only the diagram and question, measuring visual geometric reasoning without additional structural cues. Phase~B augments the diagram with the structured description, while Phase~C removes the image and retains only the structured description together with the question. Because Phase~B and Phase~C contain identical textual information, comparing these two conditions isolates the contribution of the visual modality beyond the structured representation.

\begin{table}[ht]
\centering
\caption{Three-phase evaluation protocol.}
\label{tab:phases}
\footnotesize
\resizebox{\columnwidth}{!}{%
\begin{tabular}{c l p{3.0cm}}
\toprule
\textbf{Phase} & \textbf{Input} & \textbf{Purpose} \\
\midrule
A & Diagram + Question & Diagram-only reasoning \\
B & Diagram + Desc. + Question & Diagram-plus-description reasoning \\
C & Description + Question & Description-only reasoning \\
\bottomrule
\end{tabular}%
}
\end{table}

\subsection{Experimental Settings}
\label{ssec:models}

We evaluate three open-weight VLMs (Qwen3-VL-8B-Instruct \citep{bai2025qwen3}, Gemma-4-26B \citep{team2026gemma}, and LLaMA-3.2-11B) together with two proprietary models (Gemini 2.5 Flash and GPT-4o-mini), representing a diverse set of contemporary multilingual vision-language systems; full checkpoint and parameter details are listed in Appendix~\ref{app:extended_stats} (Table~\ref{tab:models}). All models are evaluated in the zero-shot setting using identical prompts, deterministic decoding (temperature $=0$), and a maximum generation length of 64 tokens.

Predictions are evaluated using a two-stage scoring pipeline. Symbolic mathematical answers are first normalized and compared using SymPy. Predictions that cannot be symbolically matched are then evaluated by GPT-5.4-nano \citep{openai2026gpt54nano} as an LLM judge for semantic equivalence. To avoid bias from the same vendor, GPT-4o-mini outputs are judged using Gemini 2.5 Flash rather than an OpenAI model.

To determine whether Phase~B and Phase~C produce statistically equivalent performance, we perform Two One-Sided Tests (TOST) with $\alpha=0.05$ and an equivalence margin of $\pm5$\%.

\section{Benchmark Validation}
\label{sec:results}

\subsection{Evaluation on ChitraMiti (1k Test Split)}
\label{ssec:chitramiti_results}

We first evaluate whether the proposed three-phase protocol distinguishes the contribution of visual and textual information when solving Bengali geometry problems. If the structured description preserves the information required for reasoning, the models should achieve a similar performance in Phase~B (diagram-plus-description) and Phase~C (description-only). In contrast, lower performance in Phase~A (diagram-only) would indicate that current vision-language models struggle to recover the same information directly from the diagram.

Table~\ref{tab:chitramiti_results} reports the accuracy of all five models evaluated in the ChitraMiti-1k test split. In every model, Phase~B and Phase~C produce closely matched performance, whereas Phase~A remains lower for most models. For example, Gemma-4-26B achieves 34.0\% in both Phase B and Phase C (vs.\ 28.7\% in Phase A), while Qwen3-VL-8B-Instruct scores 29.8\% in Phase B and 29.1\% in Phase C (vs.\ 27.0\% in Phase A).

Phase~A is generally lower than Phase~B and Phase~C, suggesting that the structured descriptions provide information that is not always recovered from the diagram alone. Because the gaps are modest for several models, we interpret this as a consistent but limited benefit of structured descriptions rather than evidence that visual reasoning is unnecessary.

\begin{table}[t]
\centering\small
\begin{tabular}{lccc}
\toprule
Model & A & B & C \\
\midrule
Qwen3-VL-8B-Instruct & 27.0 & 29.8 & 29.1 \\
Gemma-4-26B & 28.7 & 34.0 & 34.0 \\
LLaMA-3.2-11B & 13.8 & 16.2 & 17.5 \\
Gemini 2.5 Flash & 26.5 & 29.9 & 28.4 \\
GPT-4o-mini & 16.8 & 19.2 & 21.2 \\
\bottomrule
\end{tabular}
\caption{Accuracy (\%) on ChitraMiti-1k by phase. Phase~A: diagram-only; Phase~B: diagram-plus-description; Phase~C: description-only.}
\label{tab:chitramiti_results}
\end{table}

\subsection{Validation on NCTB-500}
\label{ssec:nctb_results}

We next examine whether the same evaluation pattern holds on textbook geometry problems. NCTB-500 provides a complementary evaluation setting with NCTB curriculum-aligned diagrams, natural layout variations, scanned image artifacts, and embedded Bengali text.

Table~\ref{tab:nctb_results} reports the results of the three-phase evaluation on NCTB-500 (complete bootstrap confidence intervals in Appendix~ \ref{app:stats}). The overall trend closely matches the findings on ChitraMiti-1k: Phase~B and Phase~C again achieve similar accuracy across all five models, while Phase~A remains lower for most models. The consistency of this pattern across both benchmarks suggests that the equivalence of Phase~B and Phase~C is not specific to ChitraMiti-1k.

\begin{table}[t]
\centering
\small
\begin{tabular}{lccc}
\toprule
Model & A & B & C \\
\midrule
Qwen3-VL-8B-Instruct      & 33.0 & 39.6 & 36.6 \\
Gemma-4-26B      & 35.0 & 45.2 & 42.8 \\
LLaMA-3.2-11B    & 16.8 & 21.6 & 22.0 \\
Gemini 2.5 Flash & 21.8 & 28.4 & 25.0 \\
GPT-4o-mini      & 28.6 & 28.0 & 27.4 \\
\bottomrule
\end{tabular}
\caption{Accuracy (\%) on NCTB-500 by phase. Phase~A: diagram-only; Phase~B: diagram-plus-description; Phase~C: description-only. 95\% bootstrap confidence intervals are reported in Appendix~\ref{app:stats}.}
\label{tab:nctb_results}
\end{table}

\subsection{Equivalence Between Phase B and Phase C}
\label{ssec:tost}

The results in the previous subsections show that Phase~B and Phase~C achieve highly similar accuracies across both ChitraMiti-1k and NCTB-500. While this visual similarity is encouraging, comparable accuracies alone are insufficient to conclude that the two evaluation conditions are statistically equivalent. Since the diagnostic experiments in Section~\ref{sec:diagnostics} rely on the structured description as a controlled textual representation, we first verify this assumption using a formal statistical equivalence test.

Table~\ref{tab:tost} summarizes the Phase~B$-$Phase~C differences under the two one-sided tests (TOST) with an equivalence margin of $\pm 5$\%. Across all five models and both benchmarks, the observed differences fall within this margin, and all tests reject the null hypothesis of non-equivalence ($\alpha = 0.05$). The largest difference is 3.4\% (Gemini 2.5 Flash on NCTB-500), which remains within the pre-defined margin. Exact $p$-values and the raw Phase~B and Phase~C accuracies are reported in Table~\ref{tab:tost_full}.

These results indicate that the structured description retains sufficient geometric detail to match combined-input performance, without implying that diagrams themselves are redundant: the schema captures enough information to serve as a valid textual proxy for the diagnostics that follow.

This equivalence provides the statistical foundation for the diagnostic analyses in Section~\ref{sec:diagnostics}, which use the structured description as a controlled experimental variable to test how VLMs process geometric information and whether their predictions remain grounded in the diagram.

\begin{table}[t]
\centering\small
\begin{tabular}{lrrc}
\toprule
Model & CM $\Delta$ & NCTB $\Delta$ & Equiv.\ \\
\midrule
Qwen3-VL-8B-Instruct & $+0.7$ & $+3.0$ & Yes \\
Gemma-4-26B & $0.0$ & $+2.4$ & Yes \\
LLaMA-3.2-11B & $-1.3$ & $-0.4$ & Yes \\
Gemini 2.5 Flash & $+1.5$ & $+3.4$ & Yes \\
GPT-4o-mini & $-2.0$ & $+0.6$ & Yes \\
\bottomrule
\end{tabular}
\caption{Phase~B$-$Phase~C accuracy differences (\%). All comparisons satisfy the pre-specified $\pm5$\% TOST equivalence margin ($\alpha=0.05$). CM = ChitraMiti-1k; NCTB = NCTB-500.}
\label{tab:tost}
\end{table}

\begin{table*}[t]
\centering
\footnotesize
\begin{tabular}{lccccccl}
\toprule
& \multicolumn{3}{c}{\textbf{CM}} & \multicolumn{3}{c}{\textbf{NCTB}} & \\
\cmidrule(lr){2-4} \cmidrule(lr){5-7}
\textbf{Model} & \textbf{B} & \textbf{C} & \textbf{Diff} & \textbf{B} & \textbf{C} & \textbf{Diff} & \textbf{Equiv.\ ($p$)} \\
\midrule
Qwen3-VL-8B-Instruct & 29.8 & 29.1 & $+0.7$ & 39.6 & 36.6 & $+3.0$ & $p<0.001$ \\
Gemma-4-26B & 34.0 & 34.0 & $0.0$ & 45.2 & 42.8 & $+2.4$ & $p<0.001$ \\
LLaMA-3.2-11B & 16.2 & 17.5 & $-1.3$ & 21.6 & 22.0 & $-0.4$ & $p=0.002$ \\
Gemini 2.5 Flash & 29.9 & 28.4 & $+1.5$ & 28.4 & 25.0 & $+3.4$ & $p=0.008$ \\
GPT-4o-mini & 19.2 & 21.2 & $-2.0$ & 28.0 & 27.4 & $+0.6$ & $p=0.033$ \\
\bottomrule
\end{tabular}
\caption{Full TOST equivalence statistics ($\pm5$\% margin). CM = ChitraMiti-1k; NCTB = NCTB-500.}
\label{tab:tost_full}
\end{table*}

\section{Diagnostics}
\label{sec:diagnostics}

Having established the structured description as a validated textual representation (Section~\ref{ssec:tost}), we use it to investigate how vision-language models solve Bengali geometry problems. Specifically, we address two complementary questions. First, which components of the structured description contribute most to model performance? Second, when textual descriptions conflict with the corresponding diagram, do models verify the textual information against the visual evidence? The following experiments answer these questions through controlled modifications of the structured description, while keeping the underlying geometry unchanged.

\subsection{Ablation: Relational vs.\ Numerical Content}
\label{ssec:ablation}

We first investigate which information contained in the structured description is most important for solving geometry problems. Although the complete description combines numerical quantities and spatial relationships, these two information sources play different roles during geometric reasoning. To separate their contributions, we evaluate all five models under three Phase~C conditions: the original structured description, a \emph{relational-only} version in which all numerical values are replaced with \texttt{[NUM]}, and a \emph{values-only} version in which numerical quantities are retained while spatial expressions are removed.

The results are summarized in Table~\ref{tab:ablation}. Across all five evaluated models, retaining relational information consistently produces higher accuracy than retaining numerical values alone. For example, Gemma-4-26B achieves 29.6\% under the relational-only condition compared with 26.5\% under the values-only condition, while Qwen3-VL-8B-Instruct scores 19.2\% and 16.7\%, respectively. The same trend is observed for the remaining three models, although the magnitude of the difference varies across architectures.

Spatial relationships matter more than isolated numbers for solving ChitraMiti-1k problems: numerical values still help, but their contribution is secondary to the relational structure captured in the description.

However, identifying which information models use is only part of the picture. A model may successfully exploit relational language without verifying whether those relations are visually consistent with the underlying diagram. We examine this distinction in the next experiment by introducing controlled contradictions between the textual description and the corresponding image.

\begin{table}[t]
\centering\small
\begin{tabular}{lrrr}
\toprule
Model & Full & Rel.\ & Values \\
\midrule
Qwen3-VL-8B-Instruct & 29.1 & 19.2 & 16.7 \\
Gemma-4-26B & 34.0 & 29.6 & 26.5 \\
LLaMA-3.2-11B & 17.5 & 7.5 & 5.7 \\
Gemini 2.5 Flash & 28.4 & 26.5 & 22.6 \\
GPT-4o-mini & 21.2 & 11.8 & 10.4 \\
\bottomrule
\end{tabular}
\caption{Phase~C ablation on ChitraMiti-1k (accuracy, \%). Rel.\ = numerical values replaced with \texttt{[NUM]}; Values = spatial expressions removed. The systematic gap across all five models supports the view that the structured description captures task-critical relational information.}
\label{tab:ablation}
\end{table}

\subsection{Cross-Modal Verification Under Contradictory Descriptions}
\label{ssec:adversarial}

The ablation study shows which information models use, but not whether they verify it against the diagram. A model may rely on relational descriptions without checking whether they match the image. We test this by introducing contradictions between the structured description and the original diagram.

For each problem, we modify a single spatial relation in the structured description while keeping the question, numerical values, and diagram unchanged. The resulting description is therefore grammatically correct but visually inconsistent with the underlying geometry. All evaluations are performed under Phase~B, on a 200-item stratified subset of ChitraMiti-1k, so that models receive both the original image and the manipulated description. A visually grounded model should detect the contradiction and avoid relying on the incorrect textual relation.

Table~\ref{tab:swaps} presents the accuracy under contradictory descriptions. In all five models, accuracy falls under the swap: from unperturbed Phase-B accuracies of 16.2--34.0\% to 11.5--22.5\% once the spatial relation is contradicted, a drop of 2.2--12.0\% depending on the model. Even the strongest model is misled by the swapped text on the majority of items, showing that high performance on aligned inputs does not guarantee cross-modal verification.

The pattern holds consistently across every architecture tested, open and closed alike.

\begin{table*}[t]
\centering\small
\begin{tabular}{lccccc}
\toprule
Condition / Swap Type ($N$) & Qwen3-VL-8B-Instruct & Gemma-4-26B & LLaMA-3.2-11B & Gemini 2.5 Flash & GPT-4o-mini \\
\midrule
Unperturbed Baseline (Phase~B) & 29.8\% & 34.0\% & 16.2\% & 29.9\% & 19.2\% \\
\midrule
Perp.\ $\leftrightarrow$ Parallel ($100$) & 22.0\% & 19.0\% & 14.0\% & 26.0\% & 18.0\% \\
Above $\leftrightarrow$ Below ($40$)      & 17.5\% & 32.5\% & 5.0\%  & 25.0\% & 15.0\% \\
Inside $\leftrightarrow$ Outside ($24$)   & 12.5\% & 20.8\% & 12.5\% & 16.7\% & 12.5\% \\
Other Spatial Swaps ($36$)                & 19.4\% & 19.4\% & 11.1\% & 13.9\% & 19.4\% \\
\midrule
Adversarial Accuracy ($200$) & 19.5\% & 22.0\% & 11.5\% & 22.5\% & 17.0\% \\
\bottomrule
\end{tabular}
\caption{Unperturbed Phase-B baseline versus accuracy under each adversarial spatial-term swap type (accuracy, \%), on the 200-item adversarial subset of ChitraMiti-1k.}
\label{tab:swaps}
\end{table*}

\subsection{Qualitative Case Studies}
\label{ssec:qualitative}

Figure~\ref{fig:diagnostics} (Appendix~\ref{app:qualitative}) presents representative examples illustrating the quantitative findings from the preceding diagnostic experiments. The examples show three recurring behaviors: (i) structured descriptions compensate for failures in visual perception, (ii) models often fail to detect adversarial contradictions between text and diagrams, and (iii) multi-step geometric reasoning remains challenging even when both modalities are available. Additional examples are provided in Appendix~\ref{app:qualitative} (Table~\ref{tab:cases}).

\section{Error Analysis}
\label{sec:errors}
The diagnostic experiments in the previous section show how current VLMs use structured descriptions and visual information. We now examine where these models fail under standard evaluation, using Phase-A prediction patterns on the unmodified NCTB-500 benchmark together with category-level statistics from ChitraMiti-1k.

Across the 500 NCTB-500 problems, only 10.2\% are solved correctly by all five models, 65.0\% are missed by every model, and 24.8\% show mixed agreement (Appendix~\ref{app:error_analysis}, Table~\ref{tab:error-overlap}), indicating that failure is driven mainly by benchmark difficulty rather than architecture-specific weaknesses. The same pattern holds once performance is stratified by difficulty (Appendix~\ref{app:error_analysis}, Table~\ref{tab:error-breakdown}): both Gemma-4-26B and Qwen3-VL-8B-Instruct do markedly better on easy and medium items, but hard problems make up 78.6\% of NCTB-500 and so dominate the overall error profile.

On ChitraMiti-1k (Appendix~\ref{app:stats}, Table~\ref{tab:stratified}), coordinate geometry, quadrilaterals, and composite multi-step problems are hardest. Composite items make up 53.8\% of the full ChitraMiti-12.8k corpus (Table~\ref{tab:extended_stats}) and remain the largest single category within the 1k test split itself (45.9\%; Table~\ref{tab:stratified}), so they contribute the largest share of errors. Failures therefore arise mainly from integrating multiple geometric relationships rather than from isolated perception errors, which motivates the fine-tuning experiments below.

\section{Fine-Tuning}
\label{sec:finetuning}

We test whether task-specific training on ChitraMiti-12.8k narrows this gap by fine-tuning the open-weight vision-language model Qwen3.5-4B on the training split (11,231 items) using QLoRA, and then evaluating it on NCTB-500 under Phase~A. This checkpoint is separate from the zero-shot Qwen model evaluated in Section~\ref{sec:results}.

Table~\ref{tab:finetune} shows that fine-tuning increases in-distribution accuracy on ChitraMiti-1k from 12.7\% to 34.2\%, and out-of-distribution accuracy on NCTB-500 from 9.6\% to 23.6\%. About 65\% of the ChitraMiti-1k gain transfers to NCTB-500, despite the shift from clean rendered diagrams to scanned textbook pages.

\begin{table}[t]
\centering
\small
\begin{tabular}{@{}l cc c cc@{}}
\toprule
& \multicolumn{2}{c}{\textbf{ChitraMiti-1k}} & & \multicolumn{2}{c}{\textbf{NCTB-500}} \\
& \multicolumn{2}{c}{\textit{(In-Dist.)}} & & \multicolumn{2}{c}{\textit{(OOD)}} \\
\cmidrule(lr){2-3} \cmidrule(lr){5-6}
\textbf{Model} & \textbf{Base} & \textbf{FT} & & \textbf{Base} & \textbf{FT} \\
\midrule
Qwen3.5-4B & 12.7 & 34.2 & & 9.6 & 23.6 \\
\bottomrule
\end{tabular}
\caption{Phase~A accuracy (\%) for Qwen3.5-4B before and after fine-tuning on ChitraMiti-12.8k.}
\label{tab:finetune}
\end{table}

A gap remains to the best zero-shot system: fine-tuned Qwen3.5-4B reaches 23.6\% on NCTB-500, compared to 35.0\% for zero-shot Gemma-4-26B (Table~\ref{tab:nctb_results}). Fine-tuning on ChitraMiti-12.8k improves transfer to NCTB-500 but does not close the gap to the strongest zero-shot model.

\section{Discussion}
\label{sec:discussion}

ChitraMiti's contribution extends beyond introducing a new Bengali geometry benchmark. Synthetic diagrams with structured descriptions under a three-phase protocol let us disentangle reasoning from cross-modal verification, unlike conventional image-question benchmarks. By evaluating diagram-only (Phase~A), diagram-plus-description (Phase~B), and description-only (Phase~C) inputs, we find that Phase~B and Phase~C are statistically equivalent within the predefined margin, establishing the description as a reliable reference representation. This enables the diagnostic analyses in Section~\ref{sec:diagnostics}, which show that VLMs exploit explicit relations but rarely verify them against the diagram. High performance on conventional benchmarks therefore does not imply reliable visual grounding; protocols must explicitly test cross-modal verification.

ChitraMiti-12.8k and NCTB-500 play complementary roles in this framework: the synthetic benchmark supports controlled experiments, ablations, and fine-tuning, while NCTB-500 checks whether the resulting conclusions hold on Bengali textbook diagrams. That fine-tuning gains transfer from ChitraMiti to NCTB suggests ChitraMiti captures reasoning patterns rather than artifacts of its synthetic construction. More broadly, these results suggest that the evaluation of multimodal mathematical reasoning requires testing not only whether models can answer correctly, but also whether their answers remain grounded in visual evidence.

\section{Conclusion}
\label{sec:conclusion}

We presented ChitraMiti-12.8k, a large-scale synthetic benchmark for Bengali planar geometry reasoning, together with NCTB-500, a textbook-derived benchmark for evaluating generalization beyond the ChitraMiti distribution. To support a controlled analysis of multimodal reasoning, we introduced a three-phase evaluation protocol that separates diagram-only, diagram-plus-description, and description-only evaluation. Across the five evaluated models, structured descriptions serve as a statistically validated reference representation, letting us study how models actually use visual and textual information during geometric reasoning.

Using this evaluation framework, we show that current VLMs rely heavily on explicit relational descriptions but remain weak at verifying their consistency with the underlying diagram, and that fine-tuning on ChitraMiti-12.8k improves transfer to textbook problems in NCTB-500. We will publicly release both benchmarks and the associated evaluation framework to support future research on multilingual diagram reasoning and vision-language models for low-resource educational settings.

\section*{Limitations}
\label{sec:limitations}

\paragraph{Fine-tuning Scope} We fine-tune a single open-weight model (Qwen3.5-4B); whether the same gains hold for other architectures, larger models, or proprietary systems is untested due to resource constraints. 

\paragraph{Scoring reliability.} 
Predictions that cannot be parsed symbolically are judged by GPT-5.4-nano (and Gemini 2.5 Flash for GPT-4o-mini) for semantic equivalence. Absolute accuracy is low throughout, and NCTB-500's smaller size ($N=500$) widens its confidence intervals.

\paragraph{Evaluation confounds.} Base VLMs score higher on NCTB-500 than on ChitraMiti-1k (Section~\ref{sec:results}). Because NCTB material is publicly available, pretraining exposure on these textbook questions cannot be ruled out, and this could inflate Phase~C accuracy on NCTB-500 if models recognize question text rather than reason over it. ChitraMiti-1k's diagrams are synthetically generated rather than drawn from published curricular material, which limits this risk; the fact that Phase~B $\approx$ Phase~C equivalence also holds on ChitraMiti-1k suggests the equivalence result is not solely a memorization artifact. Our ablation and adversarial-swap diagnostics are evaluated on ChitraMiti-1k only.

The TOST equivalence conclusion depends on the predefined $\pm 5$\% margin; different margins could lead to different equivalence decisions, although Appendix~\ref{app:stats} reports results under a stricter $\pm 3$\% margin.

\paragraph{Cross-modal verification metric.} Our adversarial experiment measures implicit visual grounding via accuracy drops under contradictory descriptions, rather than explicit contradiction detection (e.g., forcing the model to output "Mismatch"). While a drop in accuracy strongly indicates text-overreliance, future work should design forced-choice verification tasks to explicitly measure a model's ability to flag text-image inconsistencies.

\paragraph{Single language and domain.} All experiments are restricted to Bengali planar geometry.

\section*{Ethical Considerations}
\label{sec:ethics}

\paragraph{Data provenance and release.} NCTB-500 diagrams are extracted from official NCTB Bengali textbooks for non-commercial research consistent with fair dealing exceptions (not a legal determination); we release only QA pairs and images, not original PDFs, and users should consult legal counsel before commercial use. ChitraMiti-12.8k and NCTB-500 annotations will be released on Hugging Face under CC~BY~4.0 (data cards in Appendix~\ref{app:datacard}); NCTB images are released for non-commercial research use only.

\paragraph{Human Annotation and dual-use.} This study involved a small-scale human evaluation with participants recruited through the academic and research networks of authors. Before taking part, all individuals were informed about the voluntary nature of their participation, expected time commitment and evaluation criteria.

\section*{Use of AI Assistants} AI writing assistance tools are used exclusively for the clarity of the writing. All experimental designs, ideas, validation, and conclusion were carried out entirely by the authors. The authors are fully accountable for the content's integrity and accuracy presented in this paper.

\bibliographystyle{acl_natbib}
\bibliography{references}

\appendix
\raggedbottom
\section{15-Attribute Schema: Full Specification}
\label{app:schema}

The 15-attribute schema specifies the visual and geometric properties required to represent planar geometry diagrams in natural language, spanning shapes, metric quantities, relational constraints, and spatial positioning. Table~\ref{tab:schema} defines each attribute in the schema.

\begin{table*}[t]
\centering\small
\begin{tabular}{@{}p{3.2cm}p{11.5cm}@{}}
\toprule
\textbf{Attribute} & \textbf{Description} \\
\midrule
Object Identification & Identifies all geometric shapes present (e.g., triangle, circle, line segment). \\
Vertex Labeling & Records all labeled points and their geometric roles (e.g., $A$, $B$, $O$). \\
Lengths & States all given or derivable side/segment lengths with units. \\
Angle Degrees & Records all explicitly labeled angle measures. \\
Parallel / Perp. & Notes all parallel ($\parallel$) and perpendicular ($\perp$) constraints between lines. \\
Intersections & Describes where and how lines, circles, or shapes intersect. \\
Tangents & Specifies tangent relationships between curves and lines/shapes. \\
Area Shading & Notes shaded or highlighted regions indicating the target area. \\
Embedded Symbols & Records special symbols inside the diagram (e.g., right-angle markers, tick marks). \\
Spatial Positioning & Describes relative positions of objects (e.g., inside, outside, above, left of). \\
Bounding Boxes & Records approximate spatial extents of complex compound regions. \\
Symmetry & Notes axes of symmetry or symmetric relations. \\
Concentric Relations & Describes concentric shapes or shared-center configurations. \\
Arcs / Sectors & Records arc spans, sector angles, and named arc segments. \\
Question Context & Extracts any textual question embedded in the problem statement. \\
\bottomrule
\end{tabular}
\caption{Full specification of the 15-attribute structured description schema. This is the schema used for all quantitative results and equivalence testing in the paper (Sections~\ref{sec:results}, \ref{sec:diagnostics}). The item card in Appendix~\ref{app:item_example} (Figure~\ref{fig:item_card}) illustrates a worked example.}
\label{tab:schema}
\end{table*}

\section{Extended Dataset and Model Statistics}
\label{app:extended_stats}

Near-duplicate detection used perceptual hashing (pHash) with a Hamming distance threshold of~8. Table~\ref{tab:pipeline} gives the full stage-by-stage filtering pipeline referenced in Section~\ref{ssec:chitramiti}, and Table~\ref{tab:annotation} gives the full annotation workflow. Table~\ref{tab:models} lists the evaluated model checkpoints referenced in Section~\ref{ssec:models}.

\begin{table}[t]
\centering\small
\begin{tabular}{lr}
\toprule
Stage & Remaining \\
\midrule
DeepVision-103k (raw) & 103,000 \\
Planar geometry subset & 52,000 \\
Uniform random sample & 26,000 \\
Keyword \& artifact filtering & 12,874 \\
Gemini 3.1 Flash Lite description & -- \\
Human audit sample & 200 \\
\bottomrule
\end{tabular}
\caption{ChitraMiti-12.8k construction pipeline (full stage counts for Section~\ref{ssec:chitramiti}).}
\label{tab:pipeline}
\end{table}

\begin{table}[t]
\centering\small
\begin{tabular}{l p{3.2cm}}
\toprule
Step & Detail \\
\midrule
Question translation & English $\to$ Bengali \\
Description generation & Gemini 3.1 Flash Lite, 15-attribute schema \\
Human audit sample & 200 items (3 annotators) \\
Correction rate & 3.0\% (6 of 200) \\
Description IAA & $\kappa = 0.81$ \\
Translation IAA & $\kappa = 0.78$ \\
\bottomrule
\end{tabular}
\caption{Annotation workflow, ChitraMiti-12.8k (full detail for Section~\ref{ssec:chitramiti}).}
\label{tab:annotation}
\end{table}

\begin{table}[t]
\centering\small
\begin{tabular}{lcl}
\toprule
\textbf{Model} & \textbf{Parameters} & \textbf{Type} \\
\midrule
Qwen3-VL-8B-Instruct & 8B & Open-weight \\
Gemma-4-26B & 26B (4B active) & Open-weight \\
LLaMA-3.2-11B & 11B & Open-weight \\
Gemini 2.5 Flash & -- & Closed API \\
GPT-4o-mini & -- & Closed API \\
\bottomrule
\end{tabular}
\caption{Evaluated vision-language models (full detail for Section~\ref{ssec:models}).}
\label{tab:models}
\end{table}

\section{Per-Category Results on NCTB-500}
\label{app:per_category}

Table~\ref{tab:nctb_dist} reports the category and question-type composition of NCTB-500, along with additional collection detail: pages were scanned at 300~dpi, and answers are Bengali prose averaging 164 characters. Table~\ref{tab:per_category} presents the per-category Phase A accuracy breakdown for zero-shot models. NCTB-500's categories (Table~\ref{tab:nctb_dist}) follow the NCTB curriculum's own topic labels rather than ChitraMiti-12.8k's DeepVision-derived taxonomy (Table~\ref{tab:extended_stats}); the two schemes are not intended to be mapped one-to-one, since they reflect different source curricula, and category-level comparisons in this paper are therefore made within each benchmark rather than across them.

\begin{table}[t]
\centering
\caption{NCTB-500 category and question-type distribution (500 QA pairs).}
\label{tab:nctb_dist}
\small
\begin{tabular}{lrr}
\toprule
\textbf{Category} & \textbf{Count} & \textbf{\%} \\
\midrule
Trigonometry    & 193 & 38.6 \\
General geometry & 116 & 23.2 \\
Shapes          & \phantom{0}50 & 10.0 \\
Triangles       & \phantom{0}40 & \phantom{0}8.0 \\
Circles         & \phantom{0}40 & \phantom{0}8.0 \\
Similarity      & \phantom{0}31 & \phantom{0}6.2 \\
Mensuration     & \phantom{0}30 & \phantom{0}6.0 \\
\midrule
\multicolumn{2}{l}{Question type} & \\
\midrule
Definition   & 181 & 36.2 \\
Analysis     & 169 & 33.8 \\
Application  & 150 & 30.0 \\
\bottomrule
\end{tabular}
\end{table}

\begin{table}[t]
\centering\small
\begin{tabular}{lcc}
\toprule
Category & Qwen3-VL-8B-Instruct & Gemma-4-26B \\
\midrule
Trigonometry & 32.7 & 34.1 \\
General geometry & 34.5 & 37.6 \\
Shapes & 38.0 & 41.3 \\
Triangles & 31.4 & 33.9 \\
Circles & 29.8 & 31.4 \\
Similarity & 28.7 & 29.8 \\
Mensuration & 27.8 & 31.1 \\
\bottomrule
\end{tabular}
\caption{Per-category Phase A accuracy (\%) on NCTB-500, zero-shot models (Table~\ref{tab:nctb_results}).}
\label{tab:per_category}
\end{table}

\section{Representative Dataset Item}
\label{app:item_example}

\begin{figure*}[htbp]
\centering
\begin{tcolorbox}[colback=teal!5, colframe=teal!40!black, title=ChitraMiti-12.8k Item Example, fonttitle=\bfseries\small, arc=1mm]
\begin{minipage}{0.35\textwidth}
\centering
\includegraphics[width=\linewidth]{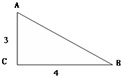}
\end{minipage}%
\hfill
\begin{minipage}{0.62\textwidth}
\raggedright\small
\textbf{Question (Bengali):} \textbengali{চিত্রে প্রদর্শিত সমকোণী ত্রিভুজ} ABC\textbengali{-কে} AC \textbengali{বাহুকে অক্ষ হিসেবে ধরে সম্পূর্ণ ঘোরালে যে ঘনবস্তু তৈরি হয়, তার আয়তন নির্ণয় করো (একক: সেমি)।} \\
\vspace{0.5em}
\textbf{Structured Description (Bengali):} \textbengali{চিত্রটিতে একটি সমকোণী ত্রিভুজ} ABC \textbengali{প্রদর্শিত হয়েছে, যার} C \textbengali{বিন্দুতে সমকোণ অবস্থিত। উল্লম্ব বাহু} AC\textbengali{-এর দৈর্ঘ্য ৩ একক এবং অনুভূমিক বাহু} CB\textbengali{-এর দৈর্ঘ্য ৪ একক। অতিভুজ} AB \textbengali{হলো ত্রিভুজটির দীর্ঘতম বাহু। ত্রিভুজটি একটি দ্বিমাত্রিক জ্যামিতিক চিত্র যা সাদা পটভূমিতে কালো রেখা দ্বারা অঙ্কিত।} \\
\vspace{0.5em}
\textbf{Gold Answer:} 50.24
\end{minipage}
\end{tcolorbox}
\caption{Representative ChitraMiti-12.8k item, showing the diagram paired with its prose description, generated by strictly following the 15-attribute schema in Table~\ref{tab:schema} without explicitly labeling the internal attributes.}
\label{fig:item_card}
\end{figure*}

Figure~\ref{fig:item_card} presents a representative item from ChitraMiti-12.8k, showing the planar geometry diagram, its 15-attribute structured Bengali description, the question, and the gold answer.

\section{System Prompts}
\label{app:prompts}

The structured description prompt below was used to generate the dataset. The evaluation prompts were used verbatim across models, with only the phase-specific input varied.

\begin{tcolorbox}[breakable, colback=teal!5, colframe=teal!40!black, title=Dataset Generation: 15-Attribute Structured Description, fonttitle=\bfseries]
\small
\ttfamily
You are an expert National Curriculum and Textbook Board geometric annotator.\\
Analyze the provided planar geometry diagram and generate a highly detailed, structured visual description in Bengali. You must strictly adhere to the following 15-attribute schema:\\
\\
1. Object identification: List all geometric shapes.\\
2. Vertex Labeling: Identify all labeled points.\\
3. Lengths: State given or derivable side lengths.\\
4. Angle Degrees: Record explicitly labeled angle measures.\\
5. Parallel / Perp.: Note parallel and perpendicular lines.\\
6. Intersections: Describe where lines or shapes intersect.\\
7. Tangents: Specify tangent relationships.\\
8. Area Shading: Note shaded target regions.\\
9. Embedded Symbols: Record special symbols (e.g. right angles).\\
10. Spatial Positioning: Describe relative positions.\\
11. Bounding Boxes: Record spatial extents of compound regions.\\
12. Symmetry: Note axes of symmetry.\\
13. Concentric Relations: Describe shared-center configurations.\\
14. Arcs / Sectors: Record arc spans and sector angles.\\
15. Question Context: Extract any embedded textual question.\\
\\
Output ONLY the final Bengali description in paragraph form, incorporating all present attributes accurately. Do not hallucinate values not explicitly marked.
\end{tcolorbox}

\begin{tcolorbox}[colback=blue!5, colframe=blue!40!black, title=Phase A: Diagram Only, fonttitle=\bfseries]
\small
\ttfamily
You are given a geometry diagram and a question in Bengali.\\
Answer the question based solely on the diagram.\\
Provide only the final answer; do not explain your reasoning.\\
{[}IMAGE{]}\\
Question: \{question\}\\
Answer:
\end{tcolorbox}

\begin{tcolorbox}[colback=green!5, colframe=green!35!black, title=Phase B: Diagram + Description, fonttitle=\bfseries]
\small
\ttfamily
You are given a geometry diagram, a structured Bengali description of the diagram, and a question in Bengali.\\
Use the diagram and description together to answer the question.\\
Provide only the final answer.\\
{[}IMAGE{]}\\
Diagram description: \{structured\_description\}\\
Question: \{question\}\\
Answer:
\end{tcolorbox}

\begin{tcolorbox}[colback=orange!5, colframe=orange!45!black, title=Phase C: Description Only, fonttitle=\bfseries]
\small
\ttfamily
You are given a structured Bengali description of a geometry diagram and a question in Bengali.\\
Answer the question based on the description.\\
Provide only the final answer.\\
Diagram description: \{structured\_description\}\\
Question: \{question\}\\
Answer:
\end{tcolorbox}

\begin{tcolorbox}[colback=purple!5, colframe=purple!40!black, title=LLM Judge, fonttitle=\bfseries]
\small
\ttfamily
You are evaluating a student's answer to a Bengali geometry question.\\
Gold answer: \{gold\}\\
Student answer: \{prediction\}\\
Are these semantically equivalent?\\
Reply with exactly one word: YES or NO.
\end{tcolorbox}

\begin{tcolorbox}[colback=gray!5, colframe=gray!50!black, title=Translation Prompt: English to Bengali, fonttitle=\bfseries]
\small
\ttfamily
System: You are an expert bilingual mathematician. Translate the geometric reasoning question from English to Bengali.\\[2pt]
Constraints: Preserve all numerical values, equations, and variable names exactly as they appear.\\[2pt]
Output: Provide ONLY the translated Bengali text.
\end{tcolorbox}

\section{Additional Statistical and Error Validation}
\label{app:stats}

\paragraph{TOST implementation.}
We treat Phase~B and Phase~C as paired item-level evaluations. For each model and benchmark, let $d_i = y_i^{B} - y_i^{C}$ for item $i$, where $y_i \in \{0,1\}$. We compute $\bar d$ and the standard error $s_d/\sqrt{n}$. With equivalence margin $\delta=0.05$, the lower one-sided test evaluates $H_{01}: \bar d \le -\delta$, and the upper one-sided test evaluates $H_{02}: \bar d \ge \delta$. Test statistics are computed as
$t_1 = (\bar d + \delta)/(s_d/\sqrt{n})$ and $t_2 = (\bar d - \delta)/(s_d/\sqrt{n})$; one-sided $p$-values are computed analytically from the Student's $t$ distribution with $n-1$ degrees of freedom using the SciPy statistical librar. No bootstrap resampling is used for the TOST $p$-values; bootstrap confidence intervals are reported separately in Table~\ref{tab:nctb_ci}.

\paragraph{NCTB-500 confidence intervals.} Table~\ref{tab:nctb_ci} reports the 95\% bootstrap confidence intervals for the NCTB-500 phase results summarized in Table~\ref{tab:nctb_results} (Section~\ref{ssec:nctb_results}).

\begin{table}[ht]
\centering
\footnotesize
\resizebox{\columnwidth}{!}{%
\begin{tabular}{lccc}
\toprule
Model & Phase A & Phase B & Phase C \\
\midrule
Qwen3-VL-8B-Instruct      & [29.0, 37.2] & [35.4, 44.0] & [32.5, 40.9] \\
Gemma-4-26B      & [30.9, 39.3] & [40.9, 49.6] & [38.5, 47.2] \\
LLaMA-3.2-11B    & [13.8, 20.3] & [18.2, 25.4] & [18.6, 25.8] \\
Gemini 2.5 Flash & [18.4, 25.6] & [24.6, 32.5] & [21.4, 29.0] \\
GPT-4o-mini      & [24.8, 32.7] & [24.2, 32.1] & [23.7, 31.5] \\
\bottomrule
\end{tabular}%
}
\caption{95\% bootstrap confidence intervals for NCTB-500 phase accuracies (Table~\ref{tab:nctb_results}).}
\label{tab:nctb_ci}
\end{table}

\paragraph{Stratified robustness.} 
Table~\ref{tab:stratified} reports Phase B and Phase C accuracy stratified by item difficulty (NCTB-500) and geometric category (ChitraMiti-1k). Note that the ChitraMiti-1k evaluation split is not proportional to the full corpus; it consists of held-out items with verified answer extraction, which changes the category mix relative to the full corpus and explains the lower composite proportion (45.9\%). The equivalence persists within the Hard (41.5 vs.\ 39.2) and Composite (27.2 vs.\ 27.7) strata, indicating that the overall agreement is not driven solely by easier examples. The Easy stratum shows a larger gap (75.0 vs.\ 66.7), likely reflecting its small sample size (approximately 12 items, 2.4\% of NCTB-500), where a few disagreements can produce substantial percentage changes. The Circle stratum (59.0 vs.\ 53.4; $\sim$234 items, 23.4\% of ChitraMiti-1k) and Coordinate Geo.\ stratum (12.0 vs.\ 17.4) slightly exceed the full-dataset $\pm5$\% margin, with Circle exhibiting the largest deviation among strata with adequate sample size.

\begin{table}[t]
\centering\small
\begin{tabular}{lccc}
\toprule
& \multicolumn{2}{c}{Gemma-4-26B} & \\
Stratum & B & C & $\Delta$ \\
\midrule
\multicolumn{4}{l}{\textit{By difficulty (NCTB-500)}} \\
Easy (2.4\%) & 75.0 & 66.7 & $+8.3$ \\
Medium (19.0\%) & 56.8 & 54.7 & $+2.1$ \\
Hard (78.6\%) & 41.5 & 39.2 & $+2.3$ \\
\multicolumn{4}{l}{\textit{By category (ChitraMiti-1k)}} \\
Circle (23.4\%) & 59.0 & 53.4 & $+5.6$ \\
Triangle (9.2\%) & 54.5 & 54.5 & $0.0$ \\
Composite (45.9\%) & 27.2 & 27.7 & $-0.5$ \\
Quadrilaterals (12.3\%) & 13.0 & 17.9 & $-4.9$ \\
Coordinate Geo.\ (9.2\%) & 12.0 & 17.4 & $-5.4$ \\
\bottomrule
\end{tabular}
\caption{Phase B vs.\ Phase C accuracy (\%), stratified by difficulty and category. Differences stay within a few points in most strata; Circle and Coordinate Geo.\ exceed the $\pm5$\% margin as discussed in the text.}
\label{tab:stratified}
\end{table}

\section{Additional Error Analysis}
\label{app:error_analysis}

Table~\ref{tab:error-overlap} provides the cross-model error agreement distribution across all 500 evaluation problems on NCTB-500. Table~\ref{tab:error-breakdown} reports the detailed performance breakdown stratified by problem difficulty for the two top-performing zero-shot models.

\begin{table}[ht]
\centering
\caption{Cross-model error agreement on NCTB-500 (Phase~A, 500 items).}
\label{tab:error-overlap}
\small
\begin{tabular}{lrr}
\toprule
\textbf{Category} & \textbf{Count} & \textbf{\%} \\
\midrule
All models correct & \phantom{0}51 & 10.2 \\
Mixed agreement    & 124 & 24.8 \\
All models wrong   & 325 & 65.0 \\
\bottomrule
\end{tabular}
\end{table}

\begin{table}[ht]
\centering
\caption{Accuracy (\%) by difficulty on NCTB-500 (Phase~A), zero-shot models.}
\label{tab:error-breakdown}
\small
\begin{tabular}{l c c}
\toprule
\textbf{Stratum} & \textbf{Qwen3-VL-8B-Instruct} & \textbf{Gemma-4-26B} \\
\midrule
\multicolumn{3}{l}{\textit{By difficulty}} \\
Hard   (78.6\%) & 31.0 & 32.8 \\
Medium (19.0\%) & 36.8 & 40.0 \\
Easy   (\phantom{1}2.4\%) & 66.7 & 66.7 \\
\bottomrule
\end{tabular}
\end{table}

\section{Additional Qualitative Case Studies}
\label{app:qualitative}

\begin{figure}[t]
\centering
\includegraphics[width=\linewidth]{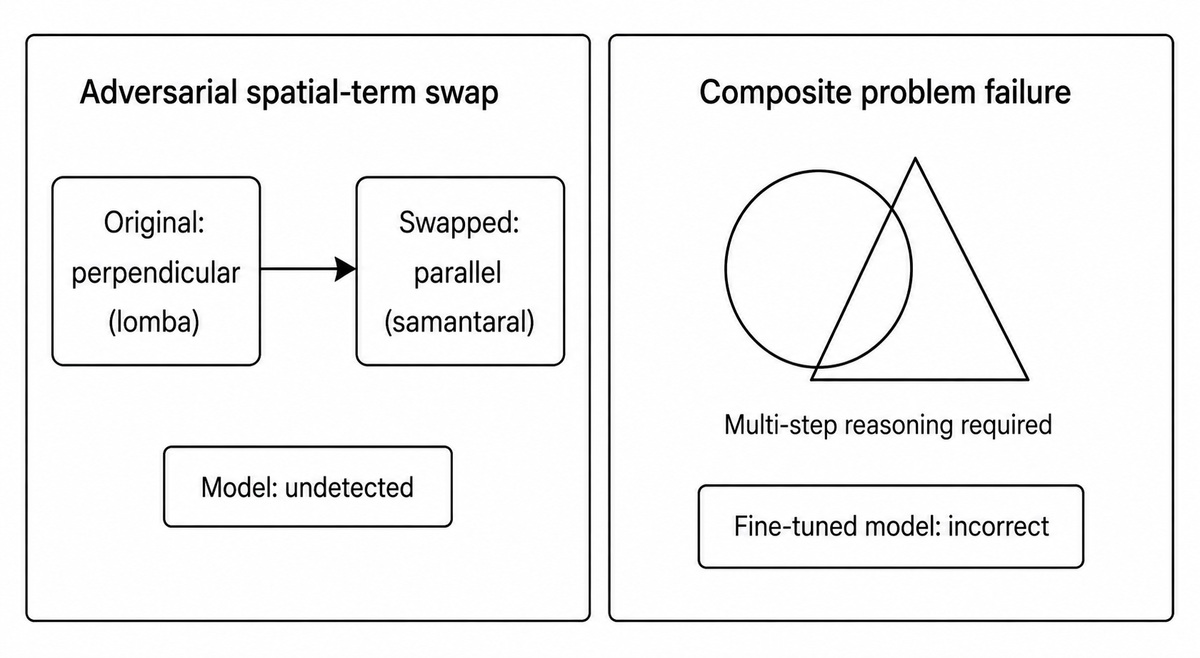}
\caption{Left: adversarial spatial-term swap, the model receives a contradictory description but does not flag the inconsistency. Right: composite problem failure, models still fail on multi-step items requiring integration of multiple geometric facts.}
\label{fig:diagnostics}
\end{figure}

Table~\ref{tab:cases} presents three representative failure modes underlying the results in Sections~\ref{sec:results} and \ref{sec:diagnostics}: text rescuing a failed visual read, unverified reliance on adversarially swapped text, and uniform failure on multi-step composite reasoning. Figure~\ref{fig:diagnostics} illustrates the adversarial swap mechanism (Section~\ref{ssec:adversarial}) alongside a representative composite failure.

\begin{table*}[t]
\centering\small
\begin{tabular}{@{}p{2.6cm}p{12.5cm}@{}}
\toprule
\textbf{Case} & \textbf{Description / Behavior} \\
\midrule
\textit{Text grounding} (Phase A fails, B/C succeeds) &
ChitraMiti (CM-4102): description states $O$ is the circle center, $AB$ a chord, $CD$ a diameter. Phase A models fail to resolve the chord intersection from raster lines alone; Phase B/C succeeds once the relation is stated explicitly. \\[4pt]
\textit{Unverified contradiction} &
ChitraMiti (CM-8819): original description states $AB \perp BC$ (gold: $90^\circ$); swapped to $AB \parallel BC$. Gemma-4-26B outputs $0^\circ$ in Phase B despite the diagram showing a clear right angle, i.e., it follows the swapped text rather than the image. \\[4pt]
\textit{Composite failure} &
NCTB-500 (NCTB-142): given $\angle BAC = 50^\circ$, find $\angle BOC$, requiring the central angle theorem plus an auxiliary triangle step. All five VLMs output $50^\circ$ or $130^\circ$ across all three phases; neither image nor text substitutes for theorem-chaining. \\
\bottomrule
\end{tabular}
\caption{Qualitative examples of the three main failure modes.}
\label{tab:cases}
\end{table*}

\section{Dataset Documentation and Data Card}
\label{app:datacard}

In accordance with data documentation guidelines, we provide data cards for ChitraMiti-12.8k and NCTB-500 below.

\begin{tcolorbox}[breakable, colback=yellow!8, colframe=yellow!45!black, title=ChitraMiti-12.8k: Data Card, fonttitle=\bfseries]
\small
\textbf{Curators:} Authors. \\
\textbf{License:} CC BY 4.0 for structured descriptions and annotations, separate from the image license described below. \\
\textbf{Source corpus:} 12,874 geometric diagrams filtered from DeepVision-103k \citep{sun2026deepvision}, whose diagram images are released under the MIT license on Hugging Face; our redistribution of these images as ChitraMiti-12.8k diagrams follows that license, separately from the CC~BY~4.0 terms covering our own descriptions and annotations. \\
\textbf{Data fields:} \texttt{image\_id}, \texttt{diagram\_image} (PNG), \texttt{bengali\_question}, \texttt{structured\_description} (15-attribute schema), \texttt{gold\_answer}, \texttt{category\_tag}. \\
\textbf{Intended use:} Benchmarking and fine-tuning multimodal foundation models for formal geometric reasoning in Bengali, with potential extension to other low-resource Indo-Aryan languages. \\
\textbf{Maintenance:} Both datasets will be hosted as versioned Hugging Face dataset repositories; corrections and additions will be tracked through the repository changelog and an accompanying issue tracker. \\
\textbf{Limitations:} Contains compound diagrams where dense topological references may require multi-step reasoning beyond current 4B to 26B model capability; see main-text Limitations for full discussion.
\end{tcolorbox}

\begin{tcolorbox}[breakable, colback=yellow!8, colframe=yellow!45!black, title=NCTB-500: Data Card, fonttitle=\bfseries]
\small
\textbf{Curators:} Three bilingual annotators with backgrounds in mathematics education; extraction and verification were coordinated by the authors. \\
\textbf{Source:} Official Bengali-medium NCTB geometry textbooks, Classes~6--10, geometry chapters. Pages were scanned at 300~dpi; diagrams were identified and cropped manually. \\
\textbf{License:} CC~BY~4.0 for derived annotations, structured descriptions, and metadata. NCTB diagram crops are released for non-commercial research use only; this is not a legal determination, and users should consult legal counsel before redistribution or commercial use. \\
\textbf{Annotation protocol:} Diagram-question pairs were extracted manually without applying OCR to diagram content. Difficulty labels were assigned during manual extraction based on cognitive complexity (e.g., single-theorem application vs.\ multi-step deduction). \\
\textbf{Data fields:} \texttt{image\_id}, \texttt{diagram\_image} (PNG crop), \texttt{bengali\_question}, \texttt{gold\_answer}, \texttt{structured\_description} (15-attribute schema), \texttt{category}, \texttt{question\_type}, \texttt{difficulty}, \texttt{grade\_level}. \\
\textbf{Intended use:} Held-out evaluation of Bengali multimodal geometric reasoning; not intended for unsupervised pretraining or commercial deployment. \\
\textbf{Maintenance:} The dataset will be hosted as a versioned Hugging Face repository; corrections will be tracked through the repository changelog and issue tracker. \\
\textbf{Limitations:} Small evaluation set; possible pretraining exposure; difficulty labels are heuristic; source textbook images are subject to source-copyright constraints.
\end{tcolorbox}

\end{document}